\documentclass{article}

 \usepackage[dblblindworkshop, final]{neurips_2025}
\workshoptitle{Multimodal Representation Learning for Healthcare}

\usepackage[utf8]{inputenc} 
\usepackage[T1]{fontenc}    
\usepackage{hyperref}       
\usepackage{url}            
\usepackage{booktabs}       
\usepackage{amsfonts}       
\usepackage{nicefrac}       
\usepackage{microtype}      
\usepackage{xcolor}         
\usepackage{amsmath}
\usepackage{graphicx}
\usepackage{subcaption}
\usepackage{comment}
\usepackage{makecell}
\title{POEMS: Product of Experts for Interpretable Multi-omic Integration using Sparse Decoding}

\author{%
Mihriban Kocak Balik$^{1}$ \quad Pekka Marttinen$^{1}$ \quad Negar Safinianaini$^{1}$ \\
$^{1}$Department of Computer Science, Aalto University, Espoo, Finland \\
\texttt{mihribannurkocak@gmail.com}, 
\texttt{\{pekka.marttinen, negar.safinianaini\}@aalto.fi}
}

\begin{document}

\maketitle

\begin{abstract}
  Integrating different molecular layers, i.e., multi-omics data, is crucial for unraveling the complexity of diseases; yet, most deep generative models either prioritize predictive performance at the expense of interpretability or enforce interpretability by linearizing the decoder, thereby weakening the network’s nonlinear expressiveness. To overcome this trade-off, we introduce POEMS: \textbf{P}roduct \textbf{O}f \textbf{E}xperts for Interpretable \textbf{M}ulti-omics Integration using \textbf{S}parse Decoding, an unsupervised probabilistic framework that preserves predictive performance while providing interpretability. POEMS provides interpretability \textit{without linearizing} any part of the network by 1) \textit{mapping} features to latent factors using sparse connections, which directly translates to biomarker discovery, 2) allowing for \textit{cross-omic associations} through a shared latent space using product of experts model, and 3) reporting \textit{contributions of each omic} by a gating network that adaptively computes their influence in the representation learning. Additionally, we present an efficient sparse decoder. In a cancer subtyping case study, POEMS achieves competitive clustering and classification performance while offering our novel set of interpretations, demonstrating that biomarker-based insight and predictive accuracy can coexist in multi-omics representation learning.

\end{abstract}

\section{Introduction}

Integrating signals from different molecular layers, i.e, omics layers (e.g., gene expression, DNA methylation, miRNA), is a key step toward understanding the complexity of biological systems and diseases. Multi-omics data provide complementary perspectives, but their high dimensionality, heterogeneity, and noise make analysis particularly challenging \cite{hasin2017multi, baiao2025technical}. Deep generative models such as variational autoencoders (VAEs) \cite{kingma2013auto, rezende2014stochastic} have become popular for representation learning in this setting. Multiple extensions have explored the combination of information across different modalities. In the context of bioinformatics, Minoura et al. \cite{minoura2021mixture} proposed scMM, a Mixture-of-Experts (MoE) based VAE for single-cell multi-omics integration, and Chen et al. \cite{chen2023mocss} developed MOCSS, an autoencoder combined with contrastive learning to allow for shared and specific representation learning for multi-omics cancer subtyping. Other approaches utilize dimensionality reduction, classical clustering algorithms, and contrastive learning to perform multi-omics integration \cite{nemo, defusion, concerto}. These approaches do not address interpretability. Those frameworks that address interpretability are typically restricted to single-omic representation learning \cite{sivae}. Moreover, even in this setting, they often enforce interpretability by linearizing the decoder network, thereby limiting the expressive power of the network \cite{ldvae,scetm,vega}. The trade-off between predictive performance and interpretability in these methods limits their broader utility in machine learning, where it is increasingly critical to understand not only whether a model performs well, but also \emph{why it learns a particular structure in the latent space and how this relates to the observed features} \cite{murdoch2019definitions, arrieta2020explainable,sidak2022interpretable}. In the context of generative models, Moran et al. \cite{moran2021identifiable} introduced the Sparse VAE, enforcing sparsity in feature-to-factor mapping (mapping features to latent factors) to promote identifiable and interpretable latent variables. However, it was designed for unimodal, not multi modal data integration. 


To address the above limitations, we introduce POEMS: \textbf{P}roduct \textbf{O}f \textbf{E}xperts for Interpretable \textbf{M}ulti-omics Integration using \textbf{S}parse Decoding, a probabilistic unsupervised representation learning framework with a novel set of interpretation tools. We achieve interpretability and strong predictive performance without linearizing the network or compromising accuracy. POEMS introduces: \textbf{1) Sparse feature-to-factor mappings} \cite{moran2021identifiable} for interpretable associations between latent factors and omics features, \textbf{2) Product-of-Experts (PoE) posterior} \cite{Hinton1999} that integrates modality-specific posteriors into a closed-form joint distribution, and \textbf{3) Gating mechanism} that adaptively weighs the contribution of each omic modality, offering a new dimension of interpretability. To scale Sparse VAEs to high-dimensional omics data, we implemented a vectorized decoder that accelerates training.

Our novel interpretability arises from: 1) \textit{sparse feature-to-factor} mappings that enable biomarker discovery, 2) \textit{cross-omic associations} captured via a shared latent space using a PoE model, and 3) \textit{adaptive per-omic contribution} estimation through a gating network. To the best of our knowledge, POEMS is the first framework to unify interpretable sparse feature–factor mappings with scalable multi-omics generative modeling in an unsupervised learning setting.

Evaluating POEMS on the cancer subtyping task using breast and kidney cancer data, i.e., BRCA and KIRC \cite{levine2013integrated}, indicates that POEMS achieves competitive cancer subtyping performance compared to the state-of-the-art methods while providing interpretable insights. These findings may contribute to the design of targeted treatments and, moreover, demonstrate that high predictive performance and interpretability can be achieved simultaneously in multi-omics unsupervised representation learning.

\section{Method}
\label{sec:method}

We propose \textbf{POEMS:} \textbf{P}roduct \textbf{O}f \textbf{E}xperts for Interpretable \textbf{M}ulti-omics Integration using \textbf{S}parse Decoding \footnote[1]{For implementation, see the \textbf{\href{https://github.com/anonymous-fish14/POEMS}{link}.}}, a probabilistic framework for interpretable representation learning from multi-omics data. POEMS constructs a \textit{shared latent space} via a Product-of-Experts (PoE) posterior and learns \textit{per-omic sparse feature-to-factor} mappings from this joint representation. This shared latent representation connects different omics modalities, enabling cross-omic interpretability. In addition, a data-dependent gating mechanism assigns modality weights that regulate \textit{each omic’s contribution} to the joint posterior. Collectively, these components yield a unified interpretable latent space that provides sparse associations between latent dimensions and omic features. To address the feature-wise decoding bottleneck of SparseVAE and make it practical for high-dimensional omics, we implement a \emph{vectorized} decoder (see Appendix~\ref{app:comp-opt} for details).


\setlength{\textfloatsep}{2pt}
\begin{figure}[h]
  \centering
\includegraphics[width=0.8\linewidth]{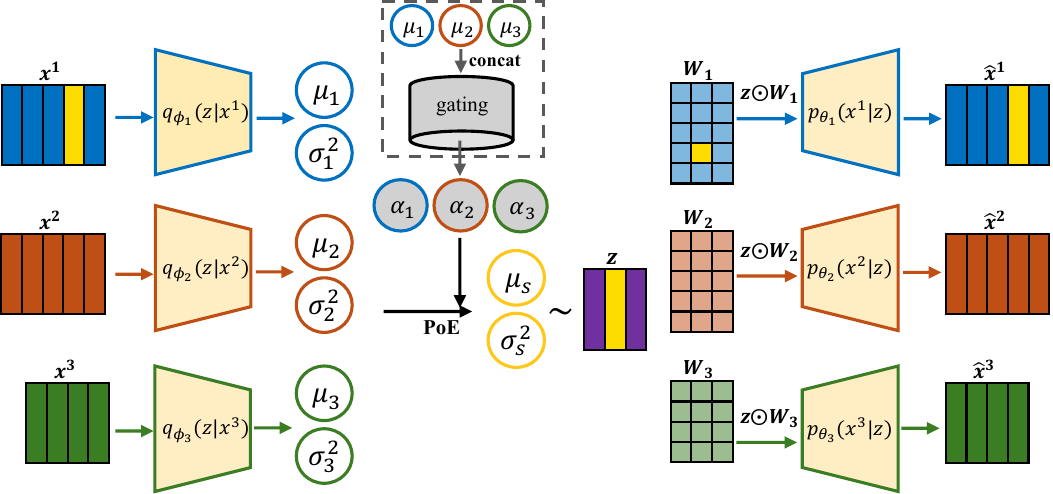}
      \caption{Schematic overview of \textbf{POEMS}. Each omics, \(\mathbf{x}^i\), is encoded into a Gaussian posterior with mean \(\boldsymbol{\mu}_i\) and variance \(\boldsymbol{\sigma}_i^2\). The posteriors are fused via a Product-of-Experts, with gating weights \(\boldsymbol{\alpha}_v\) controlling each omic’s contribution to the shared latent \(\mathbf{z}\). Before reconstruction, \(\mathbf{z}\) is modulated by the sparse feature-to-factor matrix \(\mathbf{W} \in \mathbb{R}^{D \times K}\), ensuring each feature depends on a limited subset of latent dimensions. These masked versions are then passed through modality-specific decoders for reconstruction. The yellow highlight shows the interpretable link between the 4th feature of \(\mathbf{x}^1\) and the 2nd latent factor, \(\mathbf{W}_1^{4,2}\).}
  \label{fig:poems-arch}
\end{figure}

\paragraph{Shared latent space}
Each omic \(v\!\in\!\{1,\dots,V\}\) has an encoder \(q_{\phi_v}(\mathbf{z}\mid \mathbf{x}^v)\) that outputs a Gaussian \(\mathcal{N}(\boldsymbol{\mu}_v,\boldsymbol{\sigma}_v^2)\). We fuse them with a Product-of-Experts (PoE) \cite{Hinton1999} to obtain a single inference distribution
\(q_{\phi}(\mathbf{z} \mid \mathbf{x}^{1:V}) \propto \prod_{v=1}^{V} q_{\phi_v}(\mathbf{z} \mid \mathbf{x}^v) \propto \mathcal{N}( \boldsymbol{\mu}_{s}, \boldsymbol{\sigma}_{s}^2),\)
which yields precision-weighted closed forms. To improve robustness to heterogeneous modalities and enhance interpretability, a \emph{gating network} predicts normalized weights \(\boldsymbol{\alpha}_v\!\in\![0,1]\) and rescales the precisions, as shown in Equation~\eqref{eq:poe-closed-form}. The \(\boldsymbol{\alpha}_v\) provides relative modality contributions and mitigates domination by overconfident experts.

\setlength{\parskip}{0pt}
\begin{equation}\label{eq:poe-closed-form}
\boldsymbol{\sigma}_s^2=\Big(\sum_{v=1}^{V}\boldsymbol{\alpha}_v\boldsymbol{\tau}_v\Big)^{-1},\qquad
\boldsymbol{\mu}_s=\frac{\sum_{v=1}^{V}\boldsymbol{\alpha}_v\boldsymbol{\tau}_v\,\boldsymbol{\mu}_v}{\sum_{v=1}^{V}\boldsymbol{\alpha}_v\boldsymbol{\tau}_v}, \quad \boldsymbol{\tau}_v=\boldsymbol{\sigma}_v^{-2}.
\end{equation}

\paragraph{Sparse feature-to-factor mapping} In conventional deep generative models, every latent factor is assumed to contribute to every observed feature. However, in many real-world domains such as genomics, this assumption is unrealistic: only a small subset of latent factors typically influences each feature. Sparse deep generative models address this by introducing a \emph{feature-to-factor mapping} that selectively links factors to features. Building on this idea, SparseVAE \cite{moran2021identifiable} enforces sparsity in these mappings, enabling interpretable associations between latent dimensions and meaningful subsets of features. POEMS adopts the SparseVAE approach, enforcing sparsity in the feature-to-factor mappings (\(\mathbf{W}_v\)) via a Spike-and-Slab Lasso \cite{rovckova2018} prior, yielding localized loadings \emph{connected across omics} through the shared latent variable \(\mathbf{z}\).

\paragraph{Per-omic sparse decoding} Given the shared latent \(\mathbf{z}\!\in\!\mathbb{R}^{K}\), each omic \(v\) is reconstructed via a SparseVAE-style decoder, i.e., \(p_{\theta_v}(\mathbf{x}^v\!\mid\!\mathbf{z})\), equipped with its own sparse feature-to-factor mapping \(\mathbf{W}_v\!\in\!\mathbb{R}^{D_v\times K}\). For feature \(j\) in omic \(v\), the decoder conditions on a masked latent input \(\tilde {\mathbf{z}}_{v,j}=\mathbf{z}\odot (\mathbf{W}_v)_{j}\), where \((\mathbf{W}_v)_j\) denotes \(j\)-th row of \(\mathbf{W}_v\), and \(\odot\) denotes element-wise multiplication between the latent vector and the corresponding feature–factor row. Thus, for omic \(v\), \(\mathbf{z}\odot\mathbf{W}_v\) generates all \(D_v\) masked versions of the shared latent vector simultaneously, which are passed to their corresponding decoder components for feature-wise reconstruction.

\paragraph{Objective function (ELBO)}
POEMS optimizes a single shared posterior \(q_\phi(\mathbf{z}\mid \mathbf{x}^{1:V})\). We denote \(q_\phi(\mathbf{z}\mid \mathbf{x}^{1:V})\) as \(Q_{\phi,1:V}\) and \(p_{\theta_{\nu}}(\mathbf{x}^{\nu}| \mathbf{z})\) as \(P_{\theta,\nu}\) in the ELBO, as defined in Equation~\eqref{eq:elbo}. That is, there is one KL term for the shared PoE posterior and a sum of per-omic reconstruction terms; the sparsity prior over each \(\mathbf{W}_v\) enters via the same MAP/EM-style block used in single-omic SparseVAE (mask penalty and Beta–Bernoulli updates), applied independently for each omic \(v\).
\setlength{\parskip}{0pt}
\begin{equation}\label{eq:elbo}
\sum_{v=1}^{V}\,  \Bigg\{\underbrace{\mathbb{E}_{Q_{\phi,1:V}}\!\left[\log P_{\theta,\nu}\right]}_{\text{reconstruction}}
\;+\;
\underbrace{
\mathbb{E}_{\boldsymbol{\Gamma}_v \mid \mathbf{W}_v,\, \eta_v}
\!\left[
\log p(\mathbf{W}_v \mid \boldsymbol{\Gamma}_v)\, p(\boldsymbol{\Gamma}_v \mid \eta_v)\, p(\eta_v)
\right]}_{\text{sparsity prior over } \mathbf{W}_v}\,\Bigg\}
\;-\;
\underbrace{\mathrm{KL}\!\left(Q_{\phi,1:V}\,\|\,p(\mathbf{z})\right)}_{\text{shared KL}}.
\end{equation}



\section{Experiments}
\label{sec:experiments}
We evaluate POEMS on breast and kidney cancer multi-omics datasets, i.e., BRCA and KIRC from The Cancer Genome Atlas (TCGA) \cite{levine2013integrated}, using mRNA expression, DNA methylation, and miRNA expression modalities, with details provided in Appendix~\ref{app:datasets}.POEMS is compared with representative multi-omics baselines; see Appendix~\ref{app:models}. POEMS achieves competitive or superior predictive performance across datasets. On BRCA, it attains the highest clustering and classification scores, demonstrating that combining a shared PoE-based latent space with sparsity yields discriminative representations. On KIRC, the deterministic baseline performs slightly better in clustering due to the smaller sample size, while all models reach near-perfect supervised accuracy. As this paper focuses on interpretability, detailed results of predictive performance are provided in Appendix~\ref{app:subtyping}. We examine the interpretability of POEMS on the BRCA, as we have omic labels for this dataset.

\begin{figure}[h]
  \centering
  \includegraphics[width=1.0\linewidth]{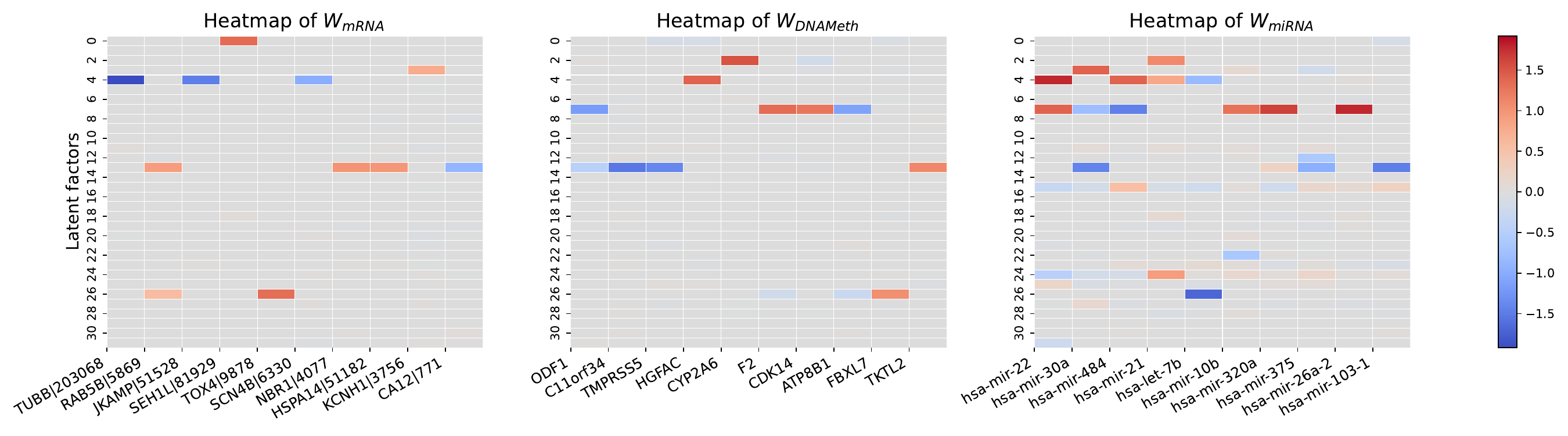}
  \caption{Top 10 activated features per omic, showing modality-specific feature-to-factor associations.}
  \label{fig:biomarker1}
\end{figure}

\paragraph{Biomarker detection} The sparse feature-to-factor mappings learned by POEMS enable biological interpretation through feature-level inspection. Figure~\ref{fig:biomarker1} shows the top 10 most influential features for each omic modality across latent dimensions. For example, latent factor 7 shows strong activation in DNA methylation and miRNA, highlighting relevant genes and regulatory patterns. Since the latent factors live in the same space among omics, one can also infer associations between omics. This finding is particularly informative, as predictive performance in multi-omics models is often dominated by mRNA, as shown in the Modality contributions section.

\setlength{\intextsep}{2.5pt}

\begin{figure}[h]
  \centering
  \includegraphics[width=1.0\linewidth]{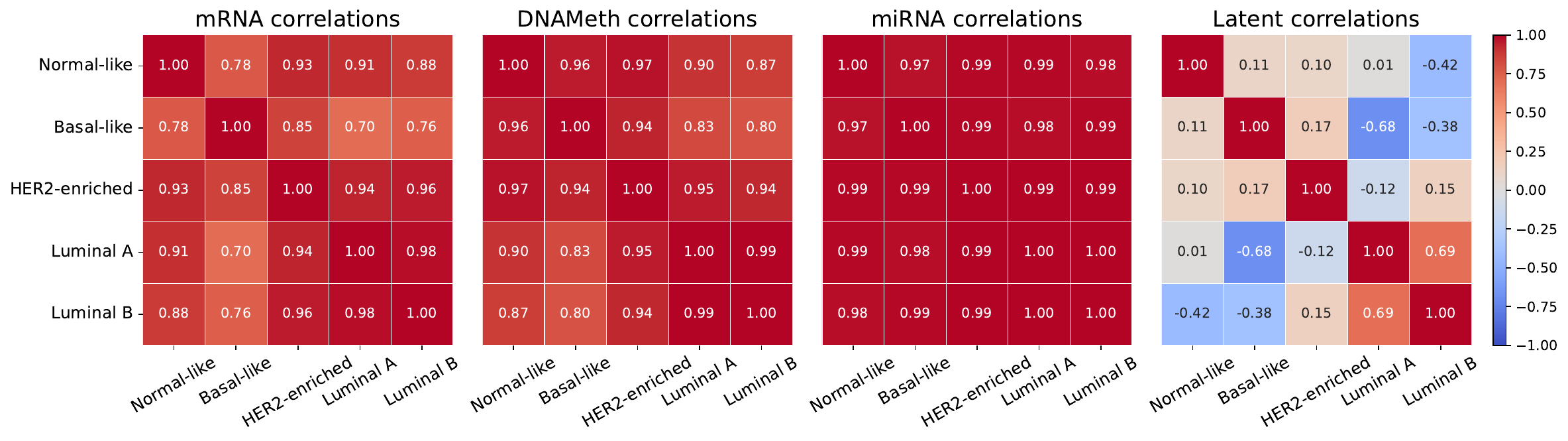}
  \caption{Subtype correlation maps with respect to input features and latent factors.}
  \label{fig:subtype-corr}
\end{figure}

\paragraph{Cancer subtype correlations}
To assess biological structure in the latent space, we compare subtype correlation matrices computed from input features and learned latent representations. As shown in Figure~\ref{fig:subtype-corr}, the latent space reveals clearer subtype relationships, such as the strong correlation between Luminal~A and Luminal~B subtypes, indicating that POEMS disentangles latent factors corresponding to clinically coherent subtypes, yielding a representation that aligns with clinical subtype structure. 

\begin{figure}[h]
  \centering
  \includegraphics[width=1.0\linewidth]{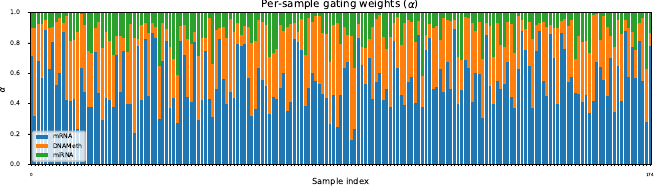}
  \caption{Per-sample gating weights (\(\alpha\)) indicating each omic’s contribution.}
  \label{fig:gating}
\end{figure}

\paragraph{Modality contributions}
The gating network in POEMS assigns modality-specific weights quantifying each omic’s contribution to the shared posterior. Figure~\ref{fig:gating} indicates that mRNA dominates overall, while DNA methylation and miRNA provide complementary signals,  showing that the model adaptively balances modalities based on their informativeness.

\setlength{\intextsep}{2.5pt}

\section{Conclusion}\label{sec:conclusion}
We presented POEMS, a sparse interpretable deep generative framework for unsupervised multi-omics integration that 1) maps features to latent factors via sparse connections for biomarker discovery, 2) captures cross-omic associations through a shared latent space using a Product-of-Experts posterior, and 3) quantifies omic-specific contributions via a gating mechanism. POEMS offers feature- and modality-level interpretability while maintaining strong cancer subtyping performance. Experiments on limited breast and kidney data reveal biologically coherent latent structures, capturing cross-omic links and modality contributions. To account for more structure in latent space, e.g., \citep{Safinianaini_Välimäki_Bresson_Gorbonos_Rajamäki_Aaltonen_Marttinen_2025}, and robustness towards hyperparameter tuning, future work can be to refine the latent space and its invariance to hyperparameter choices \citep{invariance}. Overall, POEMS demonstrates that interpretability and predictive power can coexist in deep multi-omics unsupervised learning.

\bibliographystyle{plain}
\bibliography{bibliography.bib}


\appendix

\section{Experimental setup}\label{app:experimental-setup}

\subsection{Datasets description}\label{app:datasets}
\begin{table}[h]
  \caption{Summary of multi-omics data for BRCA (breast cancer) and KIRC (kidney cancer) datasets}
  \label{table:datasets}
  \centering
  \begin{tabular}{lccccc}
    \toprule
    Dataset     & mRNA expression     & DNA methylation & miRNA expression & Samples & Subtypes \\
    \midrule
    BRCA  & 1000 & 1000 & 503 & 875 & 5 \\
    KIRC  & 58316 & 22928 & 1879 & 289 & 2 \\
    \bottomrule
  \end{tabular}
\end{table}

The evaluation of all models is conducted using two publicly available multi-omics datasets obtained from The Cancer Genome Atlas (TCGA) project~\citep{levine2013integrated}. A summary of the dataset statistics is provided in Table~\ref{table:datasets}. The first dataset, \textbf{BRCA}, represents breast cancer samples, while the second, \textbf{KIRC}, corresponds to kidney renal carcinoma. Each dataset comprises three omic layers: mRNA expression, DNA methylation, and miRNA expression profiles. The \textbf{BRCA} dataset consists of 875 samples encompassing 1,000 mRNA genes, 1,000 DNA methylation sites, and 503 miRNA features, distributed across five molecular subtypes. In contrast, the \textbf{KIRC} dataset contains 289 samples with 58,316 mRNA genes, 22,928 methylation sites, and 1,879 miRNA features, spanning two subtypes. Together, these datasets offer complementary evaluation conditions such that BRCA providing a larger and more diverse cohort suitable for examining clustering and interpretability, and KIRC serving as a smaller-scale benchmark to assess the robustness of model performance in low-sample scenarios.

\subsection{Evaluation metrics}\label{app:evaluation-metrics}

To thoroughly evaluate the learned representations, we employ both \textbf{unsupervised} and \textbf{supervised} evaluation schemes. The unsupervised metrics assess how well the latent space captures the intrinsic subtype structure, while the supervised evaluation measures how well the learned embeddings separate subtypes in a supervised setting.

\medskip
For the unsupervised evaluation, we perform \textbf{K-means clustering} on the latent embeddings and compute two complementary metrics: Normalized Mutual Information (NMI) and Accuracy (ACC). These metrics quantify the alignment between the predicted cluster assignments~\(\hat{\mathbf{y}}\) and the ground-truth subtype labels~\(\mathbf{y}\).

\begin{itemize}
    \item[\textbf{--}] \textbf{Normalized Mutual Information (NMI)} quantifies the mutual dependence between two assignments, as defined in Equation~\eqref{eq:nmi}:
    \begin{equation}\label{eq:nmi}
        \mathrm{NMI}(\mathbf{y}, \hat{\mathbf{y}}) = \frac{2 \cdot I(\mathbf{y}; \hat{\mathbf{y}})}{H(\mathbf{y}) + H(\hat{\mathbf{y}})},
    \end{equation}
    
    where \(I(\cdot;\cdot)\) denotes mutual information and \(H(\cdot)\) denotes entropy. NMI values range from~0 (no mutual information) to~1 (perfect alignment).
    
    \item[\textbf{--}] \textbf{Accuracy (ACC)} measures the proportion of correctly clustered samples after optimal label matching via the Hungarian algorithm~\citep{Kuhn1955Hungarian}, as shown in Equation~\eqref{eq:accuracy}:
    \begin{equation}\label{eq:accuracy}
        \mathrm{ACC} = \frac{1}{N} \sum_{i=1}^{N} \delta(y_i, \pi(\hat{y}_i)),
    \end{equation}
    where \(\delta(\cdot, \cdot)\) is the Kronecker delta and \(\pi\) denotes the optimal permutation mapping between cluster labels and ground truth labels.
\end{itemize}

For the supervised evaluation, we train a k-Nearest Neighbors (KNN) classifier using the latent representations and report its classification accuracy (ACC). This metric provides an additional perspective on how well the embeddings separate subtypes when label information is used, complementing the unsupervised clustering results.

\subsection{Baseline methods}\label{app:models}

To ensure a fair and comprehensive comparison, we evaluate five model variants that differ in their architectural components and training objectives but share the common goal of learning representations across multiple omic modalities:

\begin{itemize}
    \item[\textbf{--}] \textbf{MOCSS-AE}: The baseline MOCSS framework \citep{chen2023mocss}, which employs modality-specific autoencoders (AEs) to extract both shared and specific latent representations across omics.

    \item[\textbf{--}] \textbf{MOCSS-VAE}: Our modified version of MOCSS in which the autoencoders that are learning omic-specific representations are replaced by variational autoencoders (VAEs), enabling probabilistic latent inference.

    \item[\textbf{--}] \textbf{MOCSS-SparseVAE}: Our variant extends MOCSS-VAE by replacing VAEs with Sparse VAEs \citep{moran2021identifiable} for omic-specific representation learning.

    \item[\textbf{--}] \textbf{POEM}: Our variant of POEMS, replacing SparseVAEs with VAEs.

    \item[\textbf{--}] \textbf{POEMS}: Our main model proposed in this study, combining the PoE posterior with Sparse VAEs to jointly achieve interpretable, sparse feature–factor mappings and robust multi-omic integration. 
\end{itemize}

\subsection{Training configurations}\label{app:training-config}

All models were trained under a unified experimental configuration to ensure consistency and fairness in comparison. The dataset was partitioned into training, validation, and test sets following an 80\%–20\% split for training and testing, and an additional 20\% of the training portion was reserved for validation. Each model was optimized for 5{,}000 epochs with a latent dimensionality of 32, while employing early stopping to mitigate overfitting. The MOCSS-AE model (MOCSS~\citep{chen2023mocss}) followed the training procedure outlined in its original implementation, utilizing the Adam optimizer. In contrast, all remaining models adopted the AdamW optimizer. Hyperparameters (batch size (BS), learning rate (LR), and weight decay (WD)) were tuned separately for each model using a shared search grid, and the optimal configuration was determined based on the minimum total validation loss. 

\medskip
All model trainings were performed using a fixed random seed of~21 to ensure reproducibility. For the evaluation phase, K-means clustering and k-NN classification were each repeated using five different random seeds \({0, 12, 21, 42, 1234}\), and the reported subtyping results in Tables~\ref{tab:brca-results} and~\ref{tab:kirc-results} correspond to the mean and standard deviation across these runs.

\medskip
All experiments were executed on a high-performance computing (HPC) environment using a SLURM workload scheduler. Each job was run on a single CPU node with 16GB of memory and a 10-hour wall-time limit. The experiments were managed through an automated SLURM array job script that executed independent tasks corresponding to combinations of hyperparameters. The script automatically launched training runs for each configuration and stored separate output and error logs for all jobs. The exact per-model training times and total compute consumption were not systematically recorded, as computational efficiency was not the primary focus of this study. The experiments were designed to evaluate interpretability and methodological performance rather than computational scalability. Consequently, while all models were executed under identical hardware and resource constraints to ensure fair comparison, detailed runtime profiling was not required for the conclusions drawn in this work. It should be noted that the overall research project required more computational resources than the experiments directly reported in the paper. This includes additional exploratory and trial runs conducted during model development and debugging phases. These preliminary experiments were essential for refining the final methodology but are not included in the presented results.

\section{Complementary results}\label{app:results}

\subsection{Interpretability}\label{app:interpretability}

To support the interpretability analysis presented in Section~\ref{sec:experiments}, we provide additional qualitative results obtained on the test set from the BRCA dataset. These complementary visualizations show that POEMS learns biologically interpretable and structured feature–factor relationships across modalities.

\begin{figure}[h]
  \centering
  \includegraphics[width=1.0\linewidth]{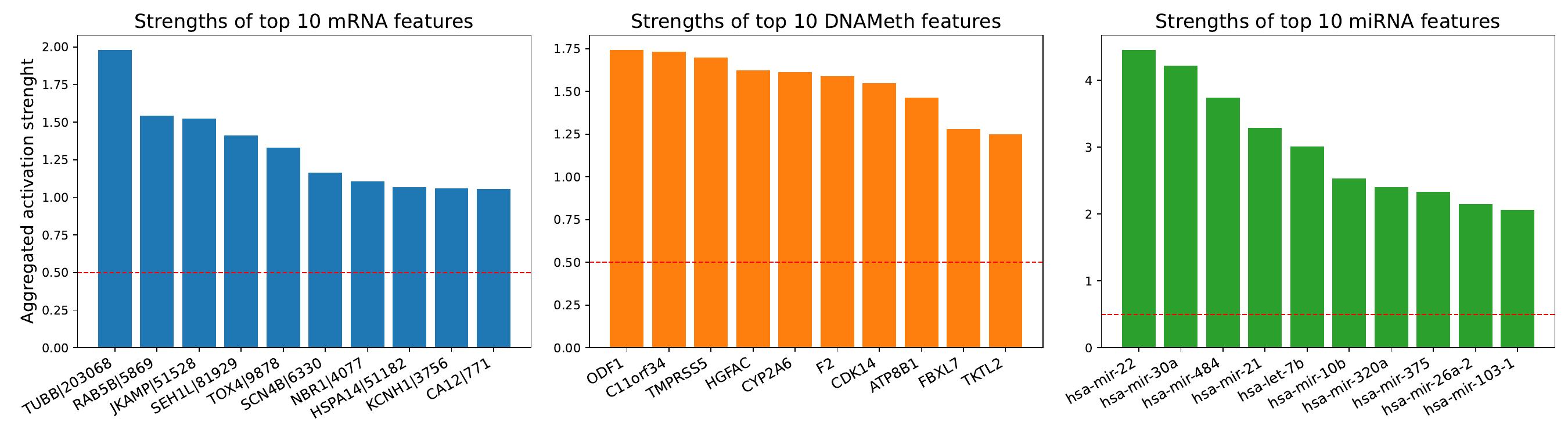}
  \caption{Aggregated activation strengths of the top 10 features across mRNA, DNAMeth, and miRNA modalities, derived from each omic’s feature–factor mapping matrix \(\mathbf{W}\). For each feature, the aggregated strength is computed as the sum of absolute loading values across all latent dimensions in \(\mathbf{W}\). The x-axis labels are the original omic feature names.}
  \label{fig:biomarker2}
\end{figure}

\medskip
Figure~\ref{fig:biomarker2} illustrates the aggregated activation strengths of the top 10 features identified within each omic (mRNA, DNAMeth, and miRNA). It highlights the dominant features associated with each latent factor, supporting the identification of shared or distinct pathways across omics. 

\begin{figure}[h]
  \centering
  \includegraphics[width=1.0\linewidth]{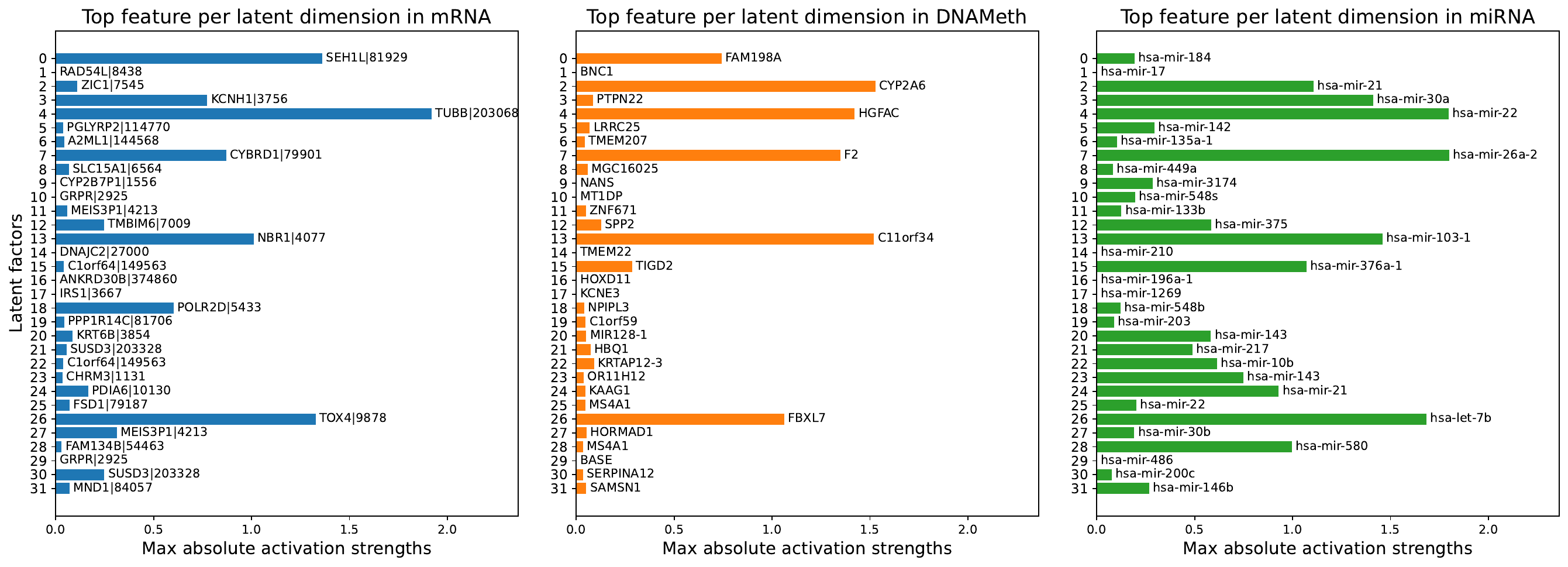}
  \caption{Top contributing features per latent dimension across mRNA, DNAMeth, and miRNA modalities. Each bar represents the maximum absolute activation strength of a feature within a given latent dimension, derived from the corresponding feature–factor mapping matrix \(\mathbf{W}\) of each omic. The feature names on the bars correspond to the most dominant input features associated with each latent factor.}
  \label{fig:biomarker3}
\end{figure}

Figure~\ref{fig:biomarker3} expands on this analysis by presenting the top contributing features per latent dimension. This mapping provides a direct interpretation of which biological variables drive particular latent directions in the representation space. 

\begin{figure}[h]
  \centering
  \includegraphics[width=1.0\linewidth]{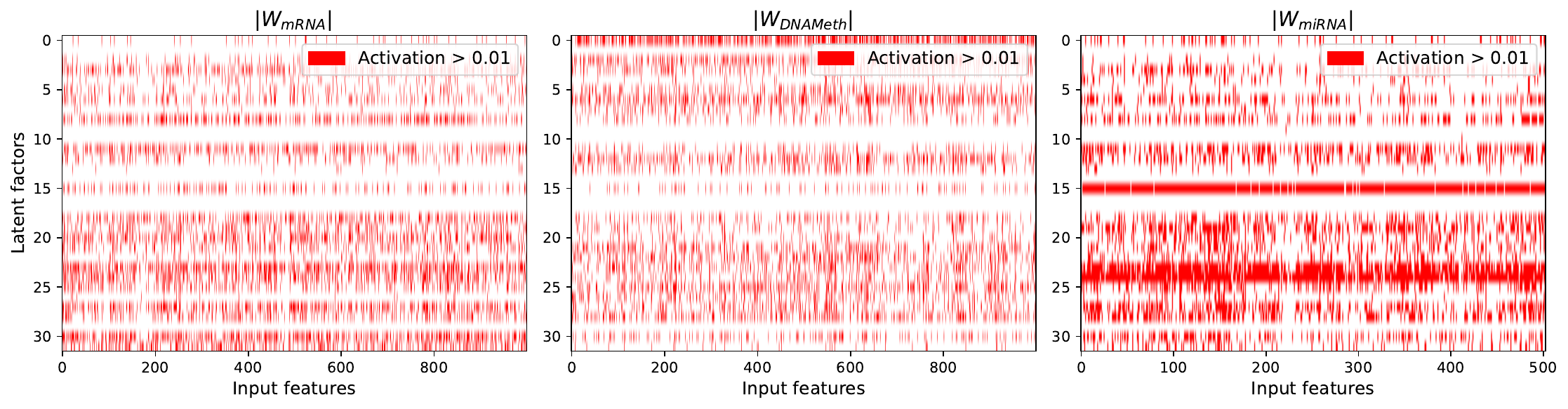}
  \caption{Binary activation maps of the absolute feature–factor loadings \(|\mathbf{W}|\) corresponding to each omic modalities mRNA, DNAMeth, and miRNA. Each row corresponds to a latent factor, and each column to an input feature. Red regions indicate active feature–factor connections with absolute activation strength greater than 0.01.}
  \label{fig:w}
\end{figure}

To visualize sparsity and activation structure within the learned weight matrices, Figure~\ref{fig:w} displays binary activation maps of the absolute loadings \(|\mathbf{W}|\) for all three omics. The resulting sparse and localized activation patterns confirm that the spike-and-slab prior effectively enforces interpretability at the feature level. 

\begin{figure}[h]
  \centering  \includegraphics[width=0.6\linewidth]{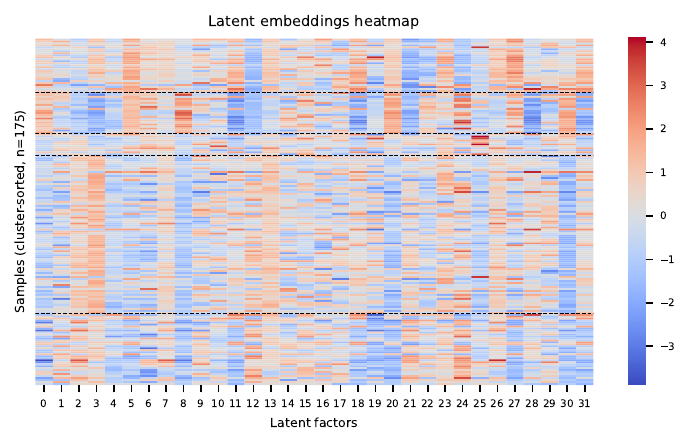}
  \caption{Heatmap of the learned latent embeddings, with samples sorted by cluster assignment. Each row represents a sample, and each column a latent factor. The dashed horizontal lines mark the boundaries between sample clusters, revealing structured variation in the latent space that reflects subtype-specific embedding patterns.}
  \label{fig:latent-embedding}
\end{figure}

Figure~\ref{fig:latent-embedding} presents the heatmap of latent embeddings, where samples are sorted according to their cluster assignments. The block-like structures along the vertical axis indicate subtype-specific activation patterns, suggesting that the latent space encodes biologically coherent sample groupings consistent with known BRCA subtypes. 

\begin{figure}[h]
  \centering
  \begin{subfigure}[b]{0.48\linewidth}
    \centering
    \includegraphics[width=\linewidth]{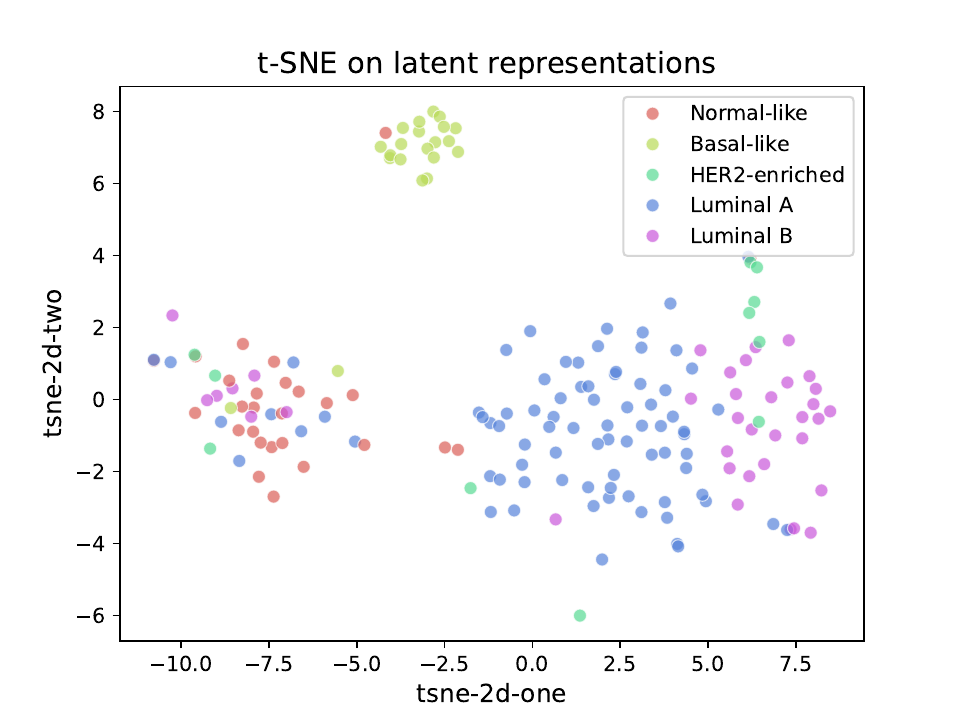}
    \caption{t-SNE projection of the latent representations.}
    \label{fig:tsne}
  \end{subfigure}
  \hfill
  \begin{subfigure}[b]{0.48\linewidth}
    \centering
    \includegraphics[width=\linewidth]{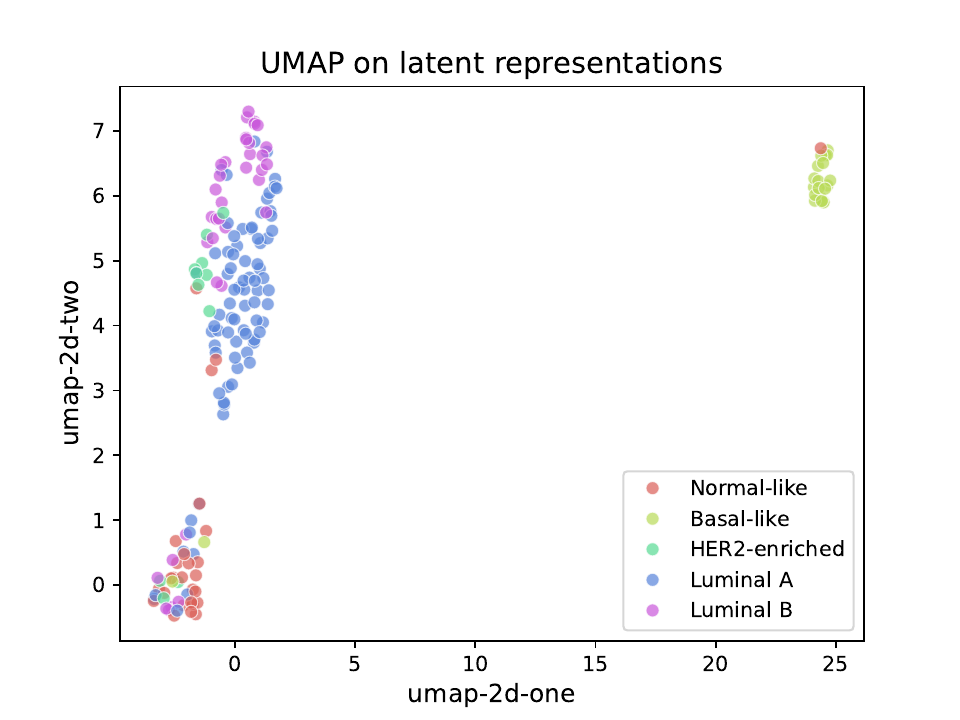}
    \caption{UMAP projection of the latent representations.}
    \label{fig:umap}
  \end{subfigure}
  \caption{Two-dimensional visualizations of the learned latent representations using t-SNE and UMAP. Each point corresponds to a sample colored by its breast cancer subtype (Normal-like, Basal-like, HER2-enriched, Luminal A, and Luminal B). Both projections reveal meaningful structure in the latent space, where samples of the same subtype form coherent clusters.}
  \label{fig:tsne-umap}
\end{figure}

Finally, Figure~\ref{fig:tsne-umap} compares two-dimensional projections of the learned latent representations using t-SNE and UMAP. Each point corresponds to a single BRCA sample colored by its molecular subtype. Both visualizations reveal meaningful clustering in the latent space, with samples of the same subtype forming separable groups. These results further confirm that the learned representations are both interpretable and discriminative with respect to relevant subtypes.

\subsection{Subtyping performance}\label{app:subtyping}

The following quantitative results complement and support the interpretability analysis presented in Section~\ref{sec:experiments}. They demonstrate that the proposed models, particularly those integrating sparsity and multi-omic fusion, not only yield latent representations that are more structured and interpretable but also achieve strong subtyping performance. These findings reinforce the conclusions drawn from the qualitative analyses, highlighting the consistency between the model’s interpretability and its discriminative capability.

\begin{table}[h]
  \caption{Subtyping performance on the BRCA dataset using K-means clustering and KNN classification applied on the resulting latent representations of the test set. Bold indicates best results.}
  \label{tab:brca-results}
  \centering
  \small
  \begin{tabular}{lccc@{\hskip 10pt}cc@{\hskip 10pt}c}
    \toprule
    Model & BS & LR & WD &
    $\text{ACC}_{\text{kmeans}}$ & $\text{NMI}_{\text{kmeans}}$ &
    $\text{ACC}_{\text{knn}}$ \\
    \midrule
    MOCSS-AE          & 256 & 3e--4 & 7e--4 & 0.62 ($\pm$0.07) & 0.41 ($\pm$0.04) & 0.68 ($\pm$0.04) \\
    MOCSS-VAE         & 512 & 5e--4 & 5e--4 & 0.57 ($\pm$0.06) & 0.43 ($\pm$0.03) & 0.73 ($\pm$0.02) \\
    MOCSS-SparseVAE   & 512 & 7e--4 & 5e--4 & 0.58 ($\pm$0.05) & 0.42 ($\pm$0.01) & 0.71 ($\pm$0.02) \\
    POEM        & 512 & 7e--4 & 3e--4 & 0.54 ($\pm$0.02) & 0.38 ($\pm$0.03) & 0.72 ($\pm$0.05) \\
    \textbf{POEMS} & 512 & 9e--4 & 1e--4 & \textbf{0.63} ($\pm$0.05) & \textbf{0.45} ($\pm$0.04) & \textbf{0.78} ($\pm$0.04) \\
    \bottomrule
  \end{tabular}
\end{table}

\medskip
As shown in Table~\ref{tab:brca-results}, \textbf{POEMS} achieves the best overall performance across all three evaluation metrics on the BRCA dataset. It attains the highest K-means clustering accuracy (0.63), NMI (0.45), and KNN classification accuracy (0.78), demonstrating that combining a PoE integration with sparsity decoding enhances both discriminative power and clustering coherence. The improvements over \textbf{MOCSS-AE} and \textbf{POEM} indicate that probabilistic inference and sparse decoding jointly contribute to the quality of the learned representations.

\begin{table}[h]
  \caption{Subtyping performance on the KIRC dataset using K-means clustering and KNN classification applied on the resulting latent representations of the test set. Bold indicates best results.}
  \label{tab:kirc-results}
  \centering
  \small
  \begin{tabular}{lccc@{\hskip 10pt}cc@{\hskip 10pt}c}
    \toprule
    Model & BS & LR & WD &
    $\text{ACC}_{\text{kmeans}}$ & $\text{NMI}_{\text{kmeans}}$ &
    $\text{ACC}_{\text{knn}}$ \\
    \midrule
    MOCSS-AE          & 32 & 3e--4 & 3e--4 & \textbf{0.93} ($\pm$0.03) & \textbf{0.60} ($\pm$0.10) & 0.99 ($\pm$0.03) \\
    MOCSS-VAE         & 32 & 5e--4 & 7e--4 & 0.91 ($\pm$0.00) & 0.54 ($\pm$0.00) & 0.99 ($\pm$0.03) \\
    MOCSS-SparseVAE   & 32 & 3e--4 & 7e--4 & 0.91 ($\pm$0.00) & 0.54 ($\pm$0.00) & \textbf{1.00} ($\pm$0.00) \\
    POEM        & 32 & 9e--4 & 9e--4 & 0.69 ($\pm$0.04) & 0.21 ($\pm$0.07) & \textbf{1.00} ($\pm$0.00) \\
    \textbf{POEMS} & 32 & 3e--4 & 1e--4 & 0.90 ($\pm$0.00) & 0.50 ($\pm$0.01) & \textbf{1.00} ($\pm$0.00) \\
    \bottomrule
  \end{tabular}
\end{table}

On the other hand, Table~\ref{tab:kirc-results} shows that on the KIRC dataset, \textbf{MOCSS-AE} outperforms all VAE-based variants in clustering metrics, achieving the highest K-means accuracy (0.93) and NMI (0.60). This suggests that deterministic architectures may be better suited for low-sample scenarios such as KIRC, where the stochasticity introduced by VAEs can lead to over-regularization. Nonetheless, all models-including \textbf{POEMS}-achieve near-perfect KNN accuracy, indicating that their latent representations remain highly discriminative when evaluated in a supervised manner.

\section{Implementation details}\label{app:implementation}
\subsection{Optimization of the Sparse VAE decoder}
\label{app:comp-opt}

As declared for the original Sparse VAE ~\citep{moran2021identifiable}, a key computational limitation of the model lies in its decoder design, which performs feature-wise reconstruction using separate masked latent vectors for each input dimension. This results in a computational complexity that scales linearly with the number of features, posing a bottleneck for high-dimensional omics data.

\medskip
To address this inefficiency, we implemented a fully vectorized version of the Sparse VAE decoder that leverages tensor broadcasting and batched operations to parallelize computations across all features simultaneously. This removes the need for sequential feature-wise decoding, significantly improving runtime efficiency. In our experiments, this optimization reduced the average training time per epoch from approximately 6.5 seconds to 1.7 seconds under identical conditions, yielding more than a threefold speedup.

\medskip
This optimization substantially enhances the scalability of Sparse VAE while preserving its interpretability benefits. Consequently, the model becomes feasible for large-scale omics applications and serves as a more practical foundation for our multi-omics extensions presented in the paper.

\end{document}